\documentclass[runningheads]{llncs}
\usepackage[T1]{fontenc}
\usepackage{graphicx}
\usepackage{newfloat}
\usepackage{listings}

\usepackage{listings}
\usepackage{graphicx}
\usepackage{subcaption}
\usepackage{multirow}
\usepackage{booktabs}
\usepackage{makecell}
\usepackage{xcolor}
\usepackage{amsmath} 
\usepackage{adjustbox}
\begin{document}
\title{Multi-dataset and Transfer Learning\\Using Gene Expression Knowledge Graphs}
\titlerunning{Multi-dataset and Transfer Learning Using Gene Expression KGs}
%
\author{Rita T. Sousa\inst{1} \and
Heiko Paulheim\inst{1}}
\authorrunning{Sousa et al.}
%
\institute{Data and web Science Group, University of Mannheim, Germany 
\email{\{rita.sousa,heiko.paulheim\}@uni-mannheim.de}}

\maketitle              
\begin{abstract}
Gene expression datasets offer insights into gene regulation mechanisms, biochemical pathways, and cellular functions. Additionally, comparing gene expression profiles between disease and control patients can deepen the understanding of disease pathology. Therefore, machine learning has been used to process gene expression data, with patient diagnosis emerging as one of the most popular applications. Although gene expression data can provide valuable insights, challenges arise because the number of patients in expression datasets is usually limited, and the data from different datasets with different gene expressions cannot be easily combined. 
This work proposes a novel methodology to address these challenges by integrating multiple gene expression datasets and domain-specific knowledge using knowledge graphs, a unique tool for biomedical data integration. Then, vector representations are produced using knowledge graph embedding techniques, which are used as inputs for a graph neural network and a multi-layer perceptron. We evaluate the efficacy of our methodology in three settings: single-dataset learning, multi-dataset learning, and transfer learning. The experimental results show that combining gene expression datasets and domain-specific knowledge improves patient diagnosis in all three settings.

\keywords{Knowledge Graph \and Knowledge Graph Embeddings \and Gene expression data \and Patient Diagnosis.}
\end{abstract}
%

\section{Introduction}

The volume of biological data being collected and accumulated is growing at an accelerated rate. 
DNA microarray chips and high-throughput sequencing are two of the latest technological innovations that allow the simultaneous measurement of the expression of different genes. 
Quantitative measurement of gene expression can offer insight into gene regulation mechanisms, biochemical pathways, and cellular functions. Furthermore, comparing gene expression profiles between patients with a disease and control patients can deepen the understanding of the pathology of the disease, facilitate the identification of genes that could serve as potential therapeutic targets~\cite{10.3389/fgene.2018.00682} or explain the response to drug treatments~\cite{chawla2022gene}. 
Given the complexity of gene expression data, it is infeasible for an expert to analyze a gene expression matrix manually. Therefore, machine learning (ML) algorithms have been used to process gene expression data, with patient diagnosis emerging as one of the more popular applications~\cite{SANDVIK2006157}.

Although gene expression datasets are easily available in public databases, and gene expression analysis is useful for biomedical applications, including patient diagnosis, processing this type of data presents some challenges~\cite{ramasamy2008key}.
While the data itself provides information on the activity levels of genes within a cell or tissue, they often do not provide a comprehensive biological context. This limitation derives from treating the expression for different genes as independent variables, an assumption that overlooks the intricate interplay and dependencies among genes.
Another challenge is the reliance of supervised ML methods on many labeled data points for effective training and performance. However, gene expression datasets typically offer only a limited number of patients, complicating the application of these methods~\cite{mansoori2018downregulation,kazerouni2020type2}.
One alternative involves combining multiple expression datasets to increase the patient pool for training ML models. However, each dataset measures gene expression across distinct genes and is measured in distinct experimental platforms, making integration difficult.

Knowledge graphs (KGs) offer a pathway towards addressing those challenges.
In biomedical systems, graphs quickly lead as the primary paradigm for modeling, learning, and reasoning~\cite{yi2022graph}. 
In a fully machine-readable format, KGs can describe real-world entities in a graph structure~\cite{hogan2021knowledge}. Additionally, KGs~\cite{rubin2008biomedical} can be enhanced by publicly accessible biomedical ontologies, which make it possible to describe domain-specific information. In the biomedical KGs, nodes represent biomedical entities or concepts, while edges illustrate their relationships. 
Over the past few years, biomedical KGs have been used in ML applications as a tool for 
data integration~\cite{nicholson2020constructing}.  

Many techniques have been proposed in ML on KGs, with graph neural networks (GNNs)~\cite{hamilton2020graph} and KG embedding methods~\cite{wang2017knowledge} becoming increasingly popular for node classification tasks.
The basic idea of GNN models is to employ a type of neural message passing in which vector messages are exchanged between nodes and updated using neural networks.
While GNN has performed very well in many tasks, it requires node features usually unavailable in biomedical real-world KGs. When node features are unavailable, GNNs often initialize them through random initialization or node statistics. Nevertheless, these approaches fail to capture a more faithful domain representation. KG embedding methods, on the contrary, present a promising avenue for capturing KG-based information since they map entities and relationships in a KG into a lower-dimensional vector space while preserving graph structure and, in some cases, semantic information.
KG embedding methods can also be used for node classification in a two-step process.
Initially, KG embedding models are employed to learn a transductive unsupervised representation of the nodes. Subsequently, these learned embeddings serve as inputs to a supervised learning algorithm, such as a Multi-Layer Perceptron (MLP).

This work tackles the challenges of using gene expression datasets for patient diagnosis. We propose a novel methodology that generates a KG to incorporate both gene expression data and publicly available datasets on genes and their functions. Then, KG embedding methods are employed to generate vector representations of KG nodes. Lastly, the vector representations of KG nodes serve as inputs for either a graph convolutional neural network (GCN) or, alternatively, into an MLP to predict the likelihood of a patient having a specific disease. 
Our methodology's efficacy is evaluated using multiple datasets for three diseases: diabetes type 2, coronary artery disease, and breast cancer.  
The proposed methodology is applied in three different scenarios: single-dataset learning, where the classifier is trained and tested with data from the same dataset; multi-dataset learning, where the classifier is trained using aggregated data from multiple datasets of the same disease and tested on one of the datasets; transfer learning, where the classifier is trained on some datasets and tested on other datasets, all focusing on the same disease.
The results demonstrated that the incorporation of domain-specific knowledge improves patient diagnosis performance, and combining data from multiple datasets is especially beneficial when the number of instances is very limited. Furthermore, the results demonstrate that our methodology can make predictions on unseen datasets during training.

In summary, the contributions of our work are as follows:
\vspace{-\topsep}
\begin{itemize}
\item An approach for building a KG that can integrate and enrich several gene expression datasets through a public dataset on gene and protein functions. 
\item An approach to tackle the lack of meaningful node features in biomedical graphs for GNNs using KG embedding methods to generate feature vectors for each node.
\item A computation-efficient method that can reuse the embedding vectors of static parts of the KG.
\item An integration of gene expression KG with supervised ML methods to support patient diagnosis.
\item An investigation of the impact of integrating multiple gene expression datasets and transferring a model trained on one dataset to another for testing. 
\end{itemize}

\section{Related Work}

Several works have explored gene expression data for patient diagnosis, employing a wide array of methodologies and datasets for a broad range of diseases, from  Alzheimer’s disease to cancer disorders~\cite{vadapalli2022artificial}.
Tan \textit{et al.}~\cite{tan2003ensemble} compare several ensemble methods for cancer classification using seven publicly available cancer-related microarray data.
Other approaches also use classical ML methods over gene expression data for cancer classification~\cite{vural2016classification,gumaei2021feature}.
Gene expression datasets related to diabetes type 2 are explored by Mansoori \textit{et al.}~\cite{mansoori2018downregulation} and Kazerouni \textit{et al.}~\cite{kazerouni2020type2}. While Kazerouni \textit{et al.}~\cite{kazerouni2020type2} compare four classical classifiers ($K$-nearest neighbor, support vector machine, logistic regression, and artificial neural networks), Mansoori \textit{et al.}~\cite{mansoori2018downregulation} use logistic regression.
Kegerreis \textit{et al.}~\cite{kegerreis2019machine} integrate gene expression data for systemic lupus erythematosus disease using several ML classifiers. Classifiers were evaluated across combined datasets or by training and testing on independent datasets.
Pirooznia \textit{et al.}~\cite{pirooznia2008comparative} present a study in which multiple ML methods, from support vector machine to random forest, are compared on how well they perform using eight gene expression datasets, each corresponding to a different disease.
Over the last few years, deep learning methods have become more prevalent and approaches that use deep learning for gene expression analysis are beginning to emerge~\cite{vadapalli2022artificial}.
Parvathavardhini \textit{et al.}~\cite{parvathavardhini2018cancer} propose a Neuro-Fuzzy approach to detect cancer by exploring gene expression data from microarray experiments. 
Shon \textit{et al.}~\cite{shon2019classification} employ a convolutional neural network algorithm for early prediction of stomach cancer. 
Schaack \textit{et al.}~\cite{schaack2021comparison} train deep-learning artificial neural networks using sepsis-related gene expression data, and test their resilience by progressively degrading the data.

These works typically rely on a singular dataset for training purposes. In some cases where multiple datasets are utilized, there is an assumption they are comparable, so they often merge them by incorporating only the shared genes as features. However, it is noted in many papers that the limited patient size of disease datasets poses a significant challenge to the predictive capability of ML models, introducing uncertainty~\cite{vadapalli2022artificial}.

\section{Methodology}

\subsection{Problem Formulation}

We tackle the challenge of handling gene expression data to support patient diagnosis.
In gene expression datasets, the expression values are organized in a matrix $n \times m$, where $n$ is the number of genes (usually more than 10 000), $m$ is the number of patients (usually more than 10), and $m<<n$. Gene expression values are numerical representations indicating the expression levels of genes under specific conditions.
The task of diagnosing patients is framed as a binary classification problem. Given a patient representation, the objective is to predict whether the patient has a specific disease (labeled as 1) or not (labeled as 0).
When using gene expression data for patient diagnosis, the patient representations correspond to rows of the $n \times m$ matrix. Most approaches use patient representations as input to a supervised ML algorithm for training prediction models~\cite{9785294,vadapalli2022artificial,10.1371/journal.pone.0230536}.
This formulation results in an oversimplification, potentially constraining these ML approaches' effectiveness and usefulness. The assumption that features are independent overlooks valuable insights about genes readily available in various biomedical resources. 
Moreover, the challenge intensifies as gene expression datasets often contain a limited number of instances, each recording expression for different genes. 
Consequently, when training prediction models, researchers either use a single dataset with little training data or attempt to merge multiple datasets. In the latter case, the datasets are typically ``incompatible'' because they have different feature sets. Thus, a naive combination usually restricts the models to using only the common features.

We propose a methodology for enriching gene expression datasets using available biomedical resources to address these challenges. We aim to do this by representing gene expression data and domain-specific knowledge in a KG, which is then fed to a supervised ML algorithm to support patient diagnosis. 
This KG not only captures relationships between genes within the same dataset, reflecting their cellular components, biological processes or molecular functions but also allows the integration of gene expression data originating from different datasets. Even when there is no direct overlap of genes across datasets, our KG is able to identify connections based on the functions the genes perform. Moreover, normalizing each patient's gene expression values mitigates discrepancies arising from varied experimental conditions. Consequently, our methodology enables seamless integration of multiple datasets for model training or transfer, even in the lack of common genes.

\subsection{Overview}

\begin{figure*}[t]
    \centering
    \includegraphics[width=\textwidth]{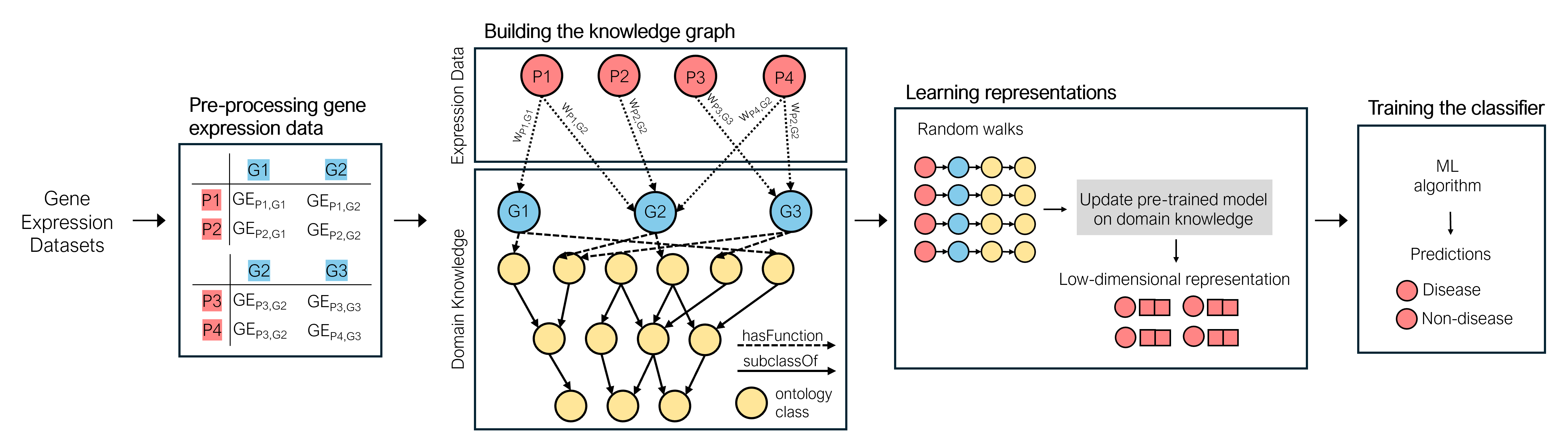}
    \caption{Overview of the proposed methodology with the main steps.}
    \label{fig:methodology}
\end{figure*}

Figure~\ref{fig:methodology} shows an overview of the methodology. The first step tackles the processing of gene expression data from experimental studies. 
The second step corresponds to building the KG, which integrates not only expression data but also domain knowledge on gene and protein functions. Representations for each node in the KG are then learned using a KG embedding method. The last step involves giving the node representations and a weighted graph as input for a GCN or, alternatively, directly feeding the node representations into an MLP. In the weighted graph, the patient's expression value for a gene is used to weigh the edge between the patient and the gene. The code for our methodology is available on GitHub (\url{https://github.com/ritatsousa/expressionKGplus}).

\subsection{Pre-processing of Gene Expression Data}

Several genomic studies have recently explored microarrays for measuring gene expression. With a single experiment, expression values are obtained simultaneously for tens of thousands of genes. The Gene Expression Omnibus (GEO)~\cite{clough2024ncbi} is a publicly accessible repository maintained by the National Center for Biotechnology Information that presents a vast collection of high-throughput gene expression and other genomics datasets. Every GEO dataset is curated, comprising biologically similar GEO samples, each representing a patient, with measurements presumed to be equivalently calculated.

The pre-processing step is essential considering the inherent complexities of microarray datasets. 
First, each probe of the microarray, identified by an identifier, contains a gene fragment for which the expression level is being determined. Each gene fragment is accompanied by an annotation detailing its biological context, indicating its association with a known gene. However, it is worth noting that not all gene fragments have such associations. Since our methodology relies on linking gene expression data with domain-specific knowledge describing gene functions, fragments without an associated gene are filtered out.
The second challenge arises from probes annotated for the same gene. The strategy adopted is, for each patient, averaging expression values across all probes corresponding to the same gene.
Additionally, expression values from different datasets may be on different scales or ranges. Therefore, we normalize the expression values of each patient using z-score normalization. Z-score normalization resizes gene expression values to align with the properties of a standard normal distribution, characterized by a mean $\mu$ of 0 and a standard deviation $\sigma$ of 1.

After the pre-processing, the data is structured in a tabular format, with each column representing a patient $P_i$, a row corresponding to a different gene $G_k$, and the cells containing gene expression values of a specific gene for a patient ${GE}_{Pi,Gk}$.

\subsection{Building the Knowledge Graph}

The KG is built by integrating two data sources: expression data and domain-specific knowledge. 
The majority of KGs are represented in RDF, a standard data model. In RDF terminology, a statement is a small piece of knowledge in the format of subject-predicate-object expressions, where the subject and the object correspond to nodes, and the predicate is the name of a relation that connects these two nodes. 

Since our methodology relies on KG graph embeddings for generating patient representations and most embedding approaches cannot handle numeric literals~\cite{preisner2022universal}, we employ a linking approach between patients and genes based on expression values. 
A link between a patient and a gene is created when the normalized patient's expression value for that gene is higher than 1 since a z-score of 1 means the data point is one standard deviation above the mean.

The domain-specific knowledge includes the Gene Ontology (GO)~\cite{GO2021} and GO annotation data~\cite{GOA2015}. The GO defines a hierarchy of classes that describes gene product functions. It can be represented as a graph where nodes are GO classes, and edges define relationships between them. The GO encompasses three domains for characterizing functions: the biological processes a gene product is involved in, the molecular functions a gene product performs, and the cellular components where a gene product is located. These three domains of GO are represented as separate root ontology classes since they do not share any common ancestor. The GO annotation data refers to assigning functions represented as GO classes to genes represented as links in the graph. 
Like most biomedical ontologies, GO is defined in OWL. Therefore, the guidelines provided by the W3C
are used to create the RDF graph. Simple axioms, including subsumption axioms and data and annotation properties, are directly converted into RDF triples. More complex axioms result in the generation of multiple triples, often requiring the use of blank nodes.

\subsection{Learning Representations}

Our methodology employs KG embedding methods to generate low-dimensional representations for each KG node.
Given the typical richness of hierarchical relations within biomedical KGs, walk-based strategies are likely better suited to capture the longer-distance relations. 
Specifically, we employ RDF2vec~\cite{ristoski2016rdf2vec}, a path-based embedding method well-suited to RDF graphs. 
RDF2vec starts by employing a depth-first search algorithm to generate random walks in a graph considering edge direction. For each node we want to learn a representation, RDF2vec generates a set number of walks with a predefined maximum depth rooted in that node.
Then, RDF2vec employs Word2vec~\cite{mikolov2013efficient}, a language model, over those walks to learn a latent low-dimensional representation of each node.

Since the domain knowledge (GO and GO annotation data) is never altered, we first train the RDF2vec model, giving it only the domain-specific KG (i.e., without patients data) as input.
Following this, for each gene expression dataset, the RDF2vec model is updated by computing new walks for the patients' dataset~\cite{hahnrdf2vec}. 
Reusing a pre-trained model circumvents the need to retrain the RDF2vec model for each dataset, thus reducing the carbon footprint and leading to faster predictions. 
Empirical experiments conducted on the same machine revealed that training the embedding model on the domain-specific KG takes approximately 9348 seconds, whereas using the pre-trained model and updating it for the patients' dataset takes only 1183 seconds. This significant reduction in running time underscores the efficiency of this approach. Additionally, it also motivates the use of RDF2vec, because such a mechanism of iteratively training KG embeddings is not supported by many methods.

\subsection{Training a Supervised Learning Algorithm}

The last step is training a supervised learning algorithm to support patient diagnosis using two distinct ML approaches. 
In the first approach, the patient embeddings are fed into an MLP classifier. 

The second approach involves giving a weighted graph as input for a GCN, using KG embeddings as node features.
In the weighted graph, the weights for edges between patients and genes represent the normalized expression values using z-score normalization. For all other edges, the weights are set to 1. We employ a multi-layer GCN architecture~\cite{Kipf2017} where embeddings are successively aggregated across several convolutional layers. 
Each convolutional layer employs a neural message passing in which vector messages are exchanged between nodes and updated using neural networks.
With each iteration of message passing, the hidden embedding $\textbf{h}_u^{(l+1)}$ associated with a node $u \in \mathcal{V}$ is updated according to the information aggregated from its neighboring nodes $\mathcal{N}(u)$. This iterative process, repeated for a fixed number of iterations, ensures that each node's final embeddings encapsulate structural and feature-based insights derived from its entire $k$-hop neighborhood. Formally, the message-passing mechanism can be expressed as outlined in~\cite{hamilton2020graph}:

\begin{equation} \label{eq:message_passing}
\textbf{h}_u^{(l+1)} = \sigma \left( \textbf{W}^{(l)} \sum_{j \in \mathcal{N}(u)} \frac{\textbf{e}_{uj}}{\sqrt{|\mathcal{N}(u)||\mathcal{N}(j)|}}\textbf{h}_j^{(l)} \right)
\end{equation}
\noindent where $\textbf{W}^{(l)}$ is a trainable parameter matrix, $\sigma$ is an activation function, and $\textbf{e}_{uj}$ is the weight on the edge from node $j$ to node $i$.
To prevent overfitting, we incorporate dropout regularization. As an activation function, we employ ReLU, while training is driven by cross-entropy loss, ensuring effective node classification. 

Transforming the KG
into an input graph for GCN results in some loss of relation information. 
The KG used to generate RDF2Vec embeddings was constructed by integrating heterogeneous relationships from the GO ontology and applying a linking strategy between patients and genes. However, the input graph for the GCN does not differentiate between edge types.
Despite this, the loss is minimized since only one edge type exists between patients and genes, and RDF2Vec embeddings used as node features still retain relational information.

\section{Evaluation}

We evaluate the proposed methodology for patient diagnosis tasks using multiple disease datasets. The following subsections describe the data used and provide an overview of the experiments.

\subsection{Data}

\begin{figure}\centering
\subfloat[Diabetes type 2]{\includegraphics[width=.33\linewidth]{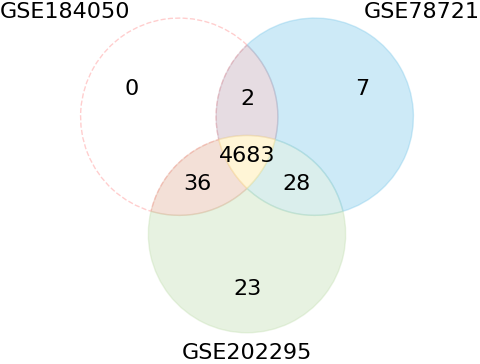}}\hfill
\subfloat[Coronary artery disease]{\includegraphics[width=.32\linewidth]{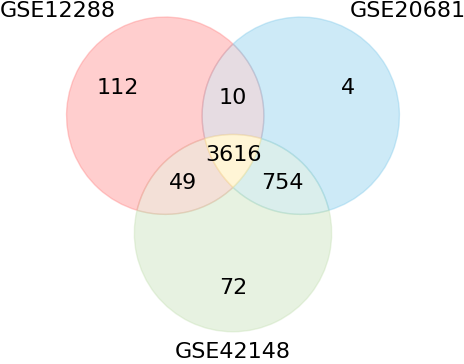}}\hfill 
\subfloat[Breast cancer]{\includegraphics[width=.32\linewidth]{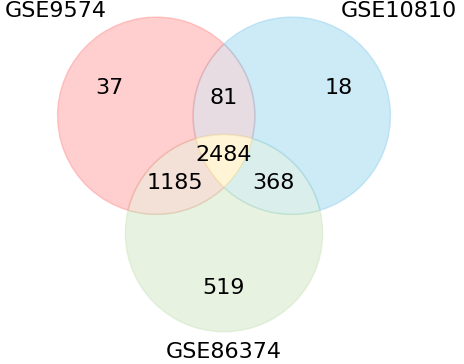}}
\caption{Venn Diagrams showing the number of genes in common between different datasets for the same disease.}
\label{fig:diagramvenn}
\end{figure}

Nine GEO datasets related to three different diseases are included in the evaluation. Each dataset comprises patients categorized into two groups: subjects diagnosed with the disease (positive examples) and those serving as control subjects (negative examples). Table~\ref{tab:statisticsdatasets} provides relevant statistics for the different datasets. For gene statistics, annotated genes encompass those with identifiers and annotations linked to biomedical ontologies. 
Additionally, Figure~\ref{fig:diagramvenn} illustrates Venn diagrams for each disease, elucidating the overlaps in the number of annotated genes across different datasets associated with the same disease. The domain KG, built by integrating GO and GO annotations, comprises 1 732 160 triples, 53 relation types, and 51 375 classes.

\begin{table}[t]
\scriptsize
\centering
\caption{Number of genes, patients, and references for each GEO dataset}
\begin{adjustbox}{width=0.42\textwidth}
\begin{tabular}{lllllll}
\toprule
\multirow{2}{*}{\textbf{Disease}} & \multirow{2}{*}{\textbf{Dataset}} & \multicolumn{2}{c}{\textbf{Genes}} && \multicolumn{2}{c}{\textbf{Patients}}  \\ \cmidrule{3-4} \cmidrule{6-7}
& & Total & Annot. && \multicolumn{1}{c}{Pos.} & \multicolumn{1}{c}{Neg.} \\ \toprule

\multirow{3}{*}{\makecell[l]{Diabetes \\ type 2}} 
        & GSE184050\footnote{https://www.ncbi.nlm.nih.gov/geo/query/acc.cgi?acc=GSE9574} & 24 737 & 4 721 && 50 & 66 \\
        & GSE78721\footnote{https://www.ncbi.nlm.nih.gov/geo/query/acc.cgi?acc=GSE78721} & 49 395 & 4 720 && 68 & 62 \\
        & GSE202295\footnote{https://www.ncbi.nlm.nih.gov/geo/query/acc.cgi?acc=GSE202295} & 66 023 & 4 770 && 61 & 50 \\ \hline
\multirow{3}{*}{\makecell[l]{Coronary \\ artery \\ disease}} 
        & GSE12288\footnote{https://www.ncbi.nlm.nih.gov/geo/query/acc.cgi?acc=GSE12288} & 22 283 & 3 787 && 110 & 112\\
        & GSE20681\footnote{https://www.ncbi.nlm.nih.gov/geo/query/acc.cgi?acc=GSE20681} & 45 015 & 4 384 && 99 & 99\\
        & GSE42148\footnote{https://www.ncbi.nlm.nih.gov/geo/query/acc.cgi?acc=GSE42148} & 62 972 & 4 491 && 13 & 11 \\ \hline
\multirow{3}{*}{\makecell[l]{Breast \\ cancer}}  
        & GSE9574\footnote{https://www.ncbi.nlm.nih.gov/geo/query/acc.cgi?acc=GSE9574} & 22 283 & 3 787 && 14 & 15 \\
        & GSE10810\footnote{https://www.ncbi.nlm.nih.gov/geo/query/acc.cgi?acc=GSE10810} & 18 382 & 2 951 && 31 & 27 \\
        & GSE86374\footnote{https://www.ncbi.nlm.nih.gov/geo/query/acc.cgi?acc=GSE86374} & 33 297 & 4 556 && 124 & 35 \\  \midrule                            
\end{tabular}
\end{adjustbox}
\label{tab:statisticsdatasets}
\end{table}

\subsection{Experimental Setup}

\begin{figure*}[t]
\centering
\subfloat[Single-dataset Learning]{\includegraphics[width=.33\linewidth]{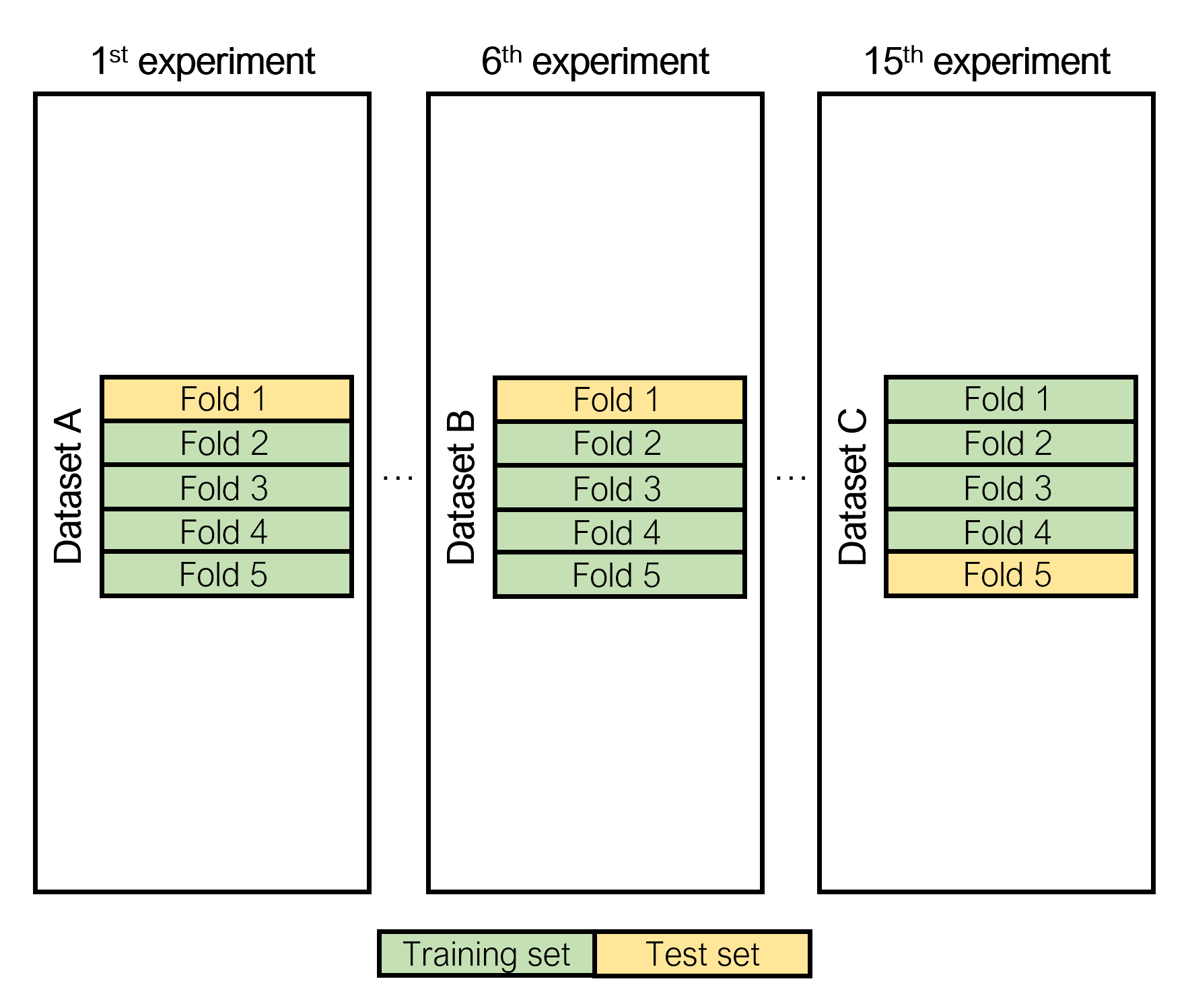}}\hfill
\subfloat[Multi-datasets Learning]{\includegraphics[width=.33\linewidth]{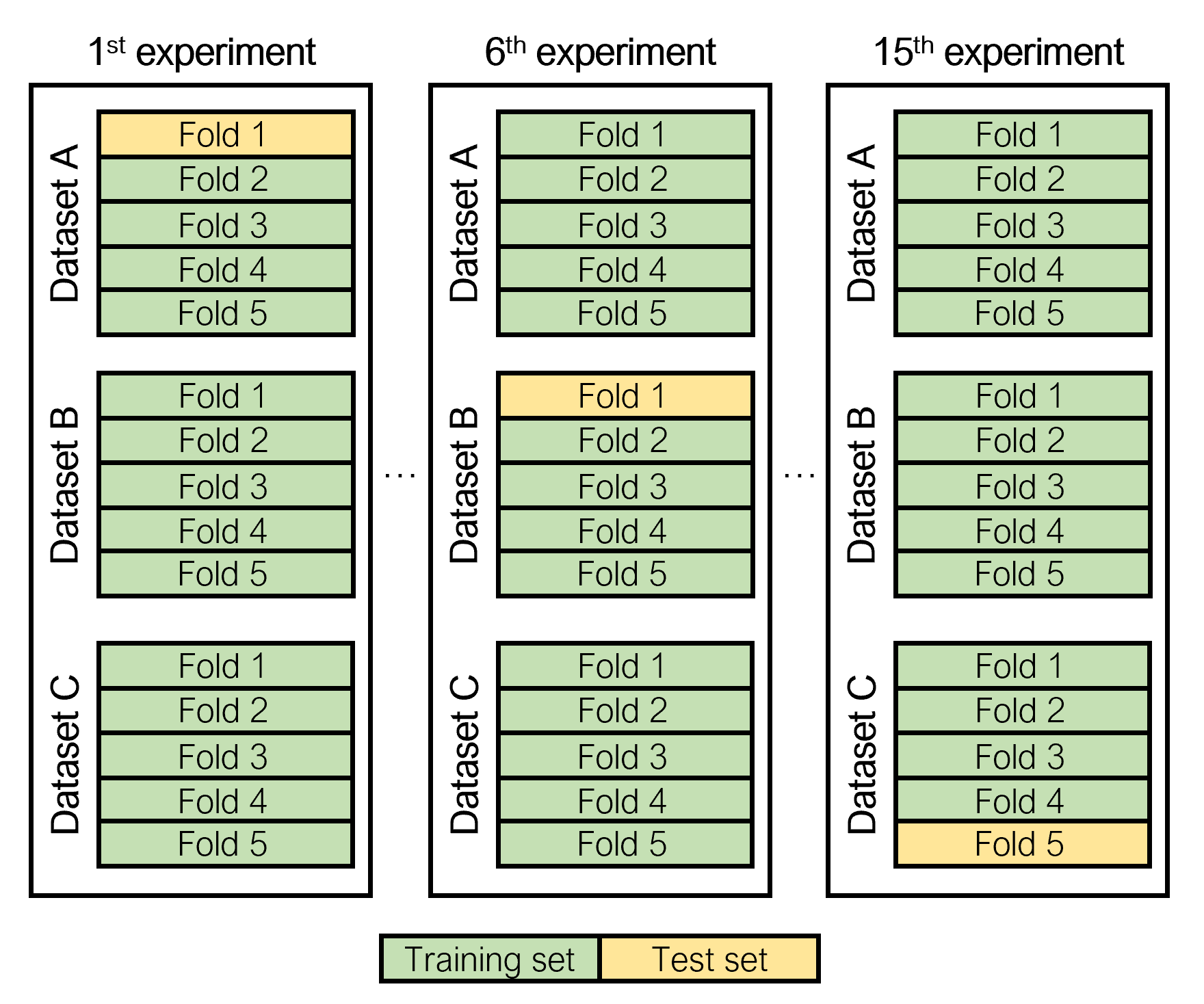}}\hfill
\subfloat[Transfer Learning]{\includegraphics[width=.33\linewidth]{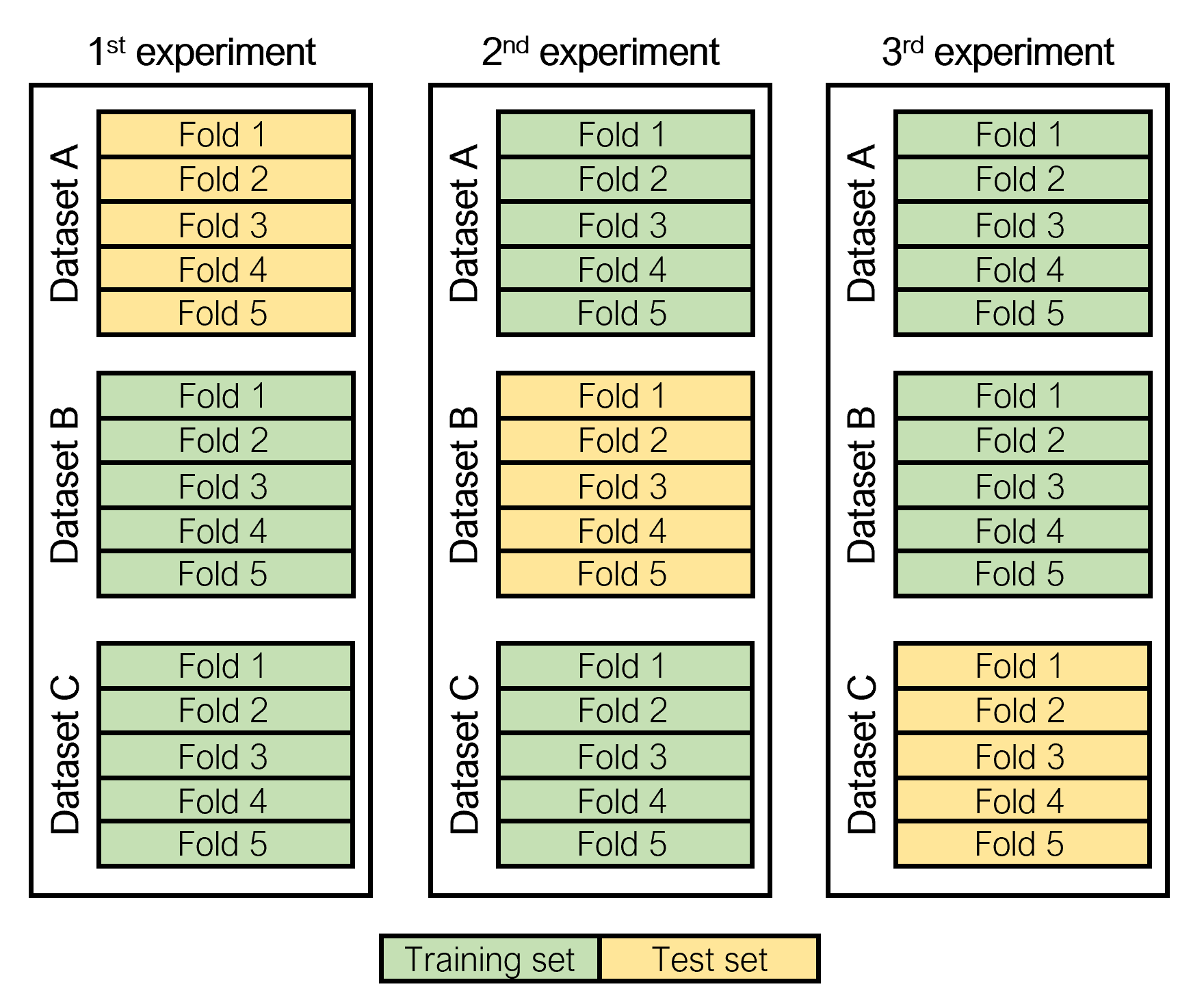}}\hfill 
\caption{Experimental strategies to split the datasets.}
\label{fig:strategy-split}
\end{figure*}

To implement our methodology, we first generate representations using RDF2vec and then train either a GCN or an MLP. All RDF2vec, MLP and GCN hyperparameters are detailed in the Appendix file.
To assess the efficacy of the proposed methodology, we analyse the classification performance using different metrics: precision, recall, F1-score, and weighted average F1-score. 
We perform stratified 5-fold cross-validation for each dataset, using the same folds throughout the different experiments. 
Our methodology enables the integration of multiple datasets, addressing the common challenge of using gene expression datasets with limited patients. Therefore, we conduct three types of experiments (Figure~\ref{fig:strategy-split}): 
\begin{itemize}
    \item \textbf{Single-dataset learning:} We only train with the data of each dataset in isolation using a 5-fold cross-validation strategy. 
    \item \textbf{Multi-datasets learning:} Since we have three datasets for each disease, we conduct experiments using combined sets from these datasets. Using the five already defined partitions of a dataset, we train the model with four partitions from one dataset and include the remaining datasets related to the same disease. 
    \item \textbf{Transfer learning:} We train the model with data from two datasets and make predictions for the third dataset.
\end{itemize}

For each setting, our methodology is compared against baselines that employ the expression values after pre-processing directly as input for an MLP. In the case of the multi-dataset and transfer learning settings, the baseline includes two variations: one that includes all genes, setting the expression value to 0 for any gene without a measured value for a given patient, and another that considers only the overlapping genes between datasets.

\section{Results and Discussion}

\begin{table}[h!]
\scriptsize
\renewcommand{\arraystretch}{1.6}
\caption{Mean and standard deviation for precision, recall, f1-score, and weighted average f1-score (Pr, Re, F1, WAF), comparing the baseline that uses gene expression values directly to our methodology when coupled with MLP or GCN for two settings (single-dataset learning and multi-dataset learning). The bold value indicates the highest performance within each setting, while the underlined value highlights the highest performance across all three settings. The performance values are marked with an asterisk (*) when the classifier predicts either label 0 or label 1 for all instances in the test set.}
\label{tab:performance}

\begin{adjustbox}{width=\textwidth}
\begin{tabular}{rllrrrrrrrrr}

\toprule
\multirow{3}{*}{\textbf{Disease}} & \multirow{3}{*}{\textbf{Dataset}} & \multirow{3}{*}{\textbf{Metric}}  & \multicolumn{3}{c}{\textbf{Single-dataset learning}}  && \multicolumn{5}{c}{\textbf{Multi-dataset learning}}  \\ \cmidrule{4-6} \cmidrule{8-12}
& & & \multicolumn{1}{c}{\multirow{2}{*}{Baseline}} & \multicolumn{2}{c}{Our methodology} & & \multicolumn{2}{c}{Baseline} & \multicolumn{3}{c}{Our methodology} \\ \cmidrule{5-6} \cmidrule{8-9} \cmidrule{11-12} 
& & & & \multicolumn{1}{c}{MLP} & \multicolumn{1}{c}{GCN} & & \multicolumn{1}{c}{All} & \multicolumn{1}{c}{Overlap} && \multicolumn{1}{c}{MLP} & \multicolumn{1}{c}{GCN} \\ \bottomrule

\multirow{12}{*}{\makecell[c]{Diabetes \\ type 2}} 
& \multirow{4}{*}{GSE184050} & Pr & 0.328 (0.298)  & 0.813 (0.157) & \textbf{\underline{0.840} (0.085)} && 0.409 (0.221) & \textbf{0.509 (0.192)} && 0.493 (0.057) & 0.504 (0.047) \\
&& Re & 0.180 (0.194) & 0.660 (0.273) & \textbf{\underline{0.740} (0.242)} && 0.480 (0.371) & 0.500 (0.261) && \textbf{0.720 (0.183)} & 0.640 (0.206) \\
&& F1 & 0.222 (0.218) & 0.675 (0.151) & \textbf{\underline{0.757} (0.138)} && 0.393 (0.226) & 0.476 (0.180) && \textbf{0.573 (0.077)} & 0.546 (0.070) \\
&& WAF & 0.495 (0.095) & 0.742 (0.084) & \textbf{\underline{0.809} (0.086)} && 0.450 (0.125) & \textbf{0.559 (0.095)} && 0.525 (0.095) & 0.525 (0.135) \\ \cmidrule{2-12}
& \multirow{4}{*}{GSE78721} & Pr & 0.428 (0.217) & 0.566 (0.085) & \textbf{\underline{0.619} (0.084)} && 0.425 (0.221) & 0.523* (0.019) && \textbf{0.511 (0.053)} & 0.456 (0.084) \\
&& Re & 0.710 (0.370) & 0.699 (0.233) & \textbf{\underline{0.780} (0.155)} && 0.552 (0.391) & 1.000* (0.000) && \textbf{0.631 (0.199)} & 0.542 (0.153) \\
&& F1 & 0.529 (0.264) & 0.614 (0.126) & \textbf{\underline{0.674} (0.038)} && 0.444 (0.264) & 0.687* (0.016) && \textbf{0.550 (0.103)} & 0.490 (0.102) \\
&& WAF & 0.391 (0.114) & 0.532 (0.113) & \textbf{\underline{0.563} (0.128)} && 0.402 (0.064) & 0.359* (0.021) && \textbf{0.452 (0.089)} & 0.401 (0.106) \\ \cmidrule{2-12}
& \multirow{4}{*}{GSE202295} & Pr & \textbf{0.695 (0.207)} & 0.581 (0.066) & 0.509 (0.109) && 0.571 (0.347) & \textbf{\underline{0.799} (0.186)} && 0.571 (0.089) & 0.452 (0.059) \\
&& Re & 0.623 (0.239) & \textbf{\underline{0.838} (0.111)} & 0.601 (0.215) && 0.326 (0.294) & \textbf{0.501 (0.322)} && 0.341 (0.129) & 0.360 (0.123) \\
&& F1 & 0.580 (0.152) & \textbf{\underline{0.680} (0.045)} & 0.547 (0.154) && 0.352 (0.234) & \textbf{0.520 (0.232)} && 0.408 (0.101) & 0.395 (0.092) \\
&& WAF & \textbf{0.507 (0.116)} & 0.504 (0.107) & 0.470 (0.114) && 0.424 (0.123) & \textbf{\underline{0.548} (0.157)} && 0.463 (0.048) & 0.407 (0.056) \\ \bottomrule

\multirow{12}{*}{\makecell[c]{Coronary \\ Artery \\ Disease}}
& \multirow{4}{*}{GSE12288} & Pr & 0.470 (0.052) & 0.553 (0.066) & \textbf{\underline{0.571} (0.064)} && 0.313 (0.256) & \textbf{0.526 (0.029)} && 0.507 (0.037) & 0.455 (0.246) \\
&& Re & 0.500 (0.246) & 0.436 (0.164) & \textbf{0.564 (0.121)} && 0.427 (0.454) & \textbf{\underline{0.664} (0.253)} && 0.518 (0.102) & 0.382 (0.224) \\
&& F1 & 0.463 (0.112) & 0.470 (0.101) & \textbf{\underline{0.562} (0.082)} && 0.329 (0.310) & \textbf{0.559 (0.104)} && 0.505 (0.038) & 0.394 (0.204) \\
&& WAF & 0.440 (0.074) & 0.523 (0.049) & \textbf{\underline{0.568} (0.075)} && 0.408 (0.079) & 0.479 (0.061) && \textbf{0.496 (0.043)} & 0.482 (0.077) \\ \cmidrule{2-12}
& \multirow{4}{*}{GSE20681} & Pr & 0.197 (0.242) & \textbf{\underline{0.557} (0.096)} & 0.543 (0.068) && 0.203 (0.248) & 0.000* (0.000) && 0.448 (0.323) & \textbf{0.536 (0.081)} \\
&& Re & 0.400 (0.490) & 0.524 (0.153) & \textbf{\underline{0.534} (0.115)} && 0.400 (0.490) & 0.000* (0.000) && 0.181 (0.196) & \textbf{0.504 (0.106)} \\
&& F1 & 0.264 (0.324) & 0.530 (0.100) & \textbf{\underline{0.535} (0.085)} && 0.269 (0.329) & 0.000* (0.000) && 0.202 (0.162) & \textbf{0.511 (0.071)} \\
&& WAF & 0.328 (0.007) & \textbf{\underline{0.544} (0.075)} & 0.542 (0.060) && 0.339 (0.007) & 0.333* (0.009) && 0.380 (0.040) & \textbf{0.520 (0.060)} \\ \cmidrule{2-12}
& \multirow{4}{*}{GSE42148} & Pr & 0.420 (0.223) & 0.470 (0.252) & \textbf{\underline{0.600} (0.374)} && 0.300 (0.400) & \textbf{0.420 (0.223)} && \textbf{0.420 (0.238)} & 0.200 (0.245) \\
&& Re & \textbf{\underline{0.800} (0.400)} & 0.733 (0.389) & 0.667 (0.365) && 0.400 (0.490) & \textbf{\underline{0.800} (0.400)} && 0.567 (0.389) & 0.200 (0.245) \\
&& F1 & 0.548 (0.282) & 0.569 (0.300) & \textbf{\underline{0.608} (0.336)} && 0.333 (0.422) & \textbf{0.548 (0.282)} && 0.477 (0.292) & 0.200 (0.245) \\
&& WAF & 0.338 (0.099) & 0.450 (0.171) & \textbf{\underline{0.564} (0.282)} && \textbf{0.448 (0.288)} & 0.338 (0.099) && 0.442 (0.207) & 0.338 (0.179) \\ \bottomrule

\multirow{12}{*}{\makecell[c]{Breast \\ Cancer}} 
& \multirow{4}{*}{GSE9574} & Pr & \textbf{0.467 (0.323)} & 0.300 (0.400) & 0.267 (0.226) && 0.450 (0.400) & 0.367 (0.371) && 0.400 (0.379) & \textbf{\underline{0.587} (0.224)} \\
&& Re & \textbf{0.567 (0.389)} & 0.267 (0.389) & 0.400 (0.389) && 0.533 (0.452) & 0.400 (0.389) && 0.533 (0.452) & \textbf{\underline{0.800} (0.267)} \\
&& F1 & \textbf{0.447 (0.245)} & 0.233 (0.291) & 0.314 (0.279) && 0.465 (0.384) & 0.360 (0.331) && 0.424 (0.355) & \textbf{\underline{0.643} (0.163)} \\
&& WAF & \textbf{0.405 (0.113)} & 0.355 (0.226) & 0.394 (0.115) && \textbf{\underline{0.578} (0.222)}  & 0.479 (0.197) && 0.394 (0.281) & 0.537 (0.188) \\ \cmidrule{2-12}
& \multirow{4}{*}{GSE10810} & Pr & 0.481 (0.419) & 0.917 (0.105) & \textbf{\underline{0.931} (0.086)} && 0.509 (0.448) & \textbf{0.800 (0.245)} && 0.776 (0.104) & 0.773 (0.104) \\
&& Re & 0.567 (0.467) & \textbf{0.905 (0.078)} & 0.838 (0.149) && 0.567 (0.467) & 0.943 (0.114) && \textbf{\underline{0.967} (0.067)} & 0.938 (0.076) \\
&& F1 & 0.508 (0.422) & \textbf{\underline{0.908} (0.083)} & 0.877 (0.109) && 0.523 (0.438) & 0.833 (0.149) && \textbf{0.855 (0.069)} & 0.844 (0.079) \\
&& WAF & 0.558 (0.293) & \textbf{\underline{0.897} (0.099)} & 0.879 (0.103) && 0.576 (0.316) & 0.700 (0.306) && 0.779 (0.179) & \textbf{0.802 (0.107)} \\ \cmidrule{2-12}
& \multirow{4}{*}{GSE86374} & Pr & 0.467 (0.382) & \textbf{\underline{0.930} (0.062)} & 0.919 (0.059) && 0.824 (0.088) & 0.624 (0.312) && \textbf{0.917 (0.032)} & 0.883 (0.036) \\
&& Re & 0.600 (0.490) & 0.892 (0.176) & \textbf{\underline{0.909} (0.163)} && 0.816 (0.368) & 0.800 (0.400) && 0.862 (0.086) & \textbf{0.878 (0.095)} \\
&& F1 & 0.525 (0.429) & 0.903 (0.122) & \textbf{\underline{0.906} (0.107)} && 0.730 (0.291) & 0.701 (0.350) && \textbf{0.885 (0.047)} & 0.876 (0.045) \\
&& WAF & 0.441 (0.296) & \textbf{\underline{0.869} (0.140)} & 0.865 (0.127) && 0.586 (0.194) & 0.562 (0.242) && \textbf{0.834 (0.054)} & 0.810 (0.051) \\ \bottomrule                            
\end{tabular}
\end{adjustbox}
\vspace{-0.4cm}
\end{table}

Our methodology has two variants, one using an MLP with RDF2Vec embeddings as input and another using a GCN with RDF2Vec embeddings as node features for patient diagnosis.
Table \ref{tab:performance} presents the performance of these two variants compared to baseline performance for single-dataset and multi-dataset settings. At the end of each dataset fold, we compute the performance metrics and report the average of the five folds.

In the single-dataset setting, using our methodology does not imply integrating data from other datasets but rather injecting domain-specific knowledge. The results confirm the hypothesis that contextualizing genetic information improves patient diagnosis, with considerable improvements for some datasets.
Notably, for the GSE184050 dataset, the f1-score increases from 0.222 in the baseline to 0.675 and 0.757 for KG embedding and GCN, respectively. The exception is the GSE9574 dataset for all metrics, the GSE41148 dataset for precision and the GSE202295 dataset for precision and f-score. These datasets are the smallest for each respective disease, which might explain why our methodology does not show improvement compared to the baseline.

In the multi-dataset setting, supervised learning models are trained with data from all datasets related to the same disease.
For this setting, a comparison between the two variations of the baselines shows that using only the gene expression data for overlapping genes generally achieves better performance. However, there are two datasets where this approach is unsuccessful, leading to a classifier that labels all test instances as either positive or negative, indicating that the model is not generalizing at all.
As in the single-dataset learning setting, for most datasets, our methodology improves the performance of baselines that use gene expression values as features.
It is worth mentioning that the baselines use as features the gene expression values, leading to a significantly higher number of features than our MLP variant that uses 100-dimensional embeddings.

When comparing the results between the multi-dataset and the single-dataset settings, it becomes evident that training with a diverse range of data sources can enhance performance in smaller datasets such as GSE9574 (consisting of 29 patients), where adding only the domain knowledge is not enough.
These improvements emphasize the importance of leveraging multiple datasets, as incorporating data from different datasets not only amplifies the amount of training data but also enriches the model's understanding.

\begin{table}[t!]
\centering
\scriptsize
\renewcommand{\arraystretch}{1.6}
\caption{Precision, recall, f1-score, and weighted average f1-score (Pr, Re, F1, WAF), comparing the baseline to our methodology when coupled with MLP or GCN for the transfer learning setting. The performance values are marked with an asterisk (*) when the classifier predicts either label 0 or label 1 for all instances in the test set.} 
\label{tab:performance2}

\begin{adjustbox}{width=0.42\textwidth}

\begin{tabular}{rllrrrrr}

\toprule
\multirow{2}{*}{\textbf{Disease}} & \multirow{2}{*}{\textbf{Dataset}} & \multirow{2}{*}{\textbf{Metric}}  &  \multicolumn{2}{c}{\textbf{Baseline}} && \multicolumn{2}{c}{\textbf{Ours}} \\ \cmidrule{4-5} \cmidrule{7-8}
& & & \multicolumn{1}{c}{All} & \multicolumn{1}{c}{Overlap} && \multicolumn{1}{c}{MLP} & \multicolumn{1}{c}{GCN}  \\ \bottomrule

\multirow{12}{*}{\makecell[c]{Diabetes \\ Type 2}}         
                                          & \multirow{4}{*}{GSE184050} & Pr  & 0.333  & \textbf{0.426}  && 0.347 & 0.342 \\
                                          &                            & Re  & 0.040  & \textbf{0.980}  && 0.500 & 0.500 \\
                                          &                            & F1  & 0.071  & \textbf{0.594}  && 0.410 & 0.407 \\
                                          &                            & WAF & \textbf{0.432}  & 0.256  && 0.373 & 0.363 \\ \cmidrule{2-8}
                                          & \multirow{4}{*}{GSE78721}  & Pr  & 0.523* & \textbf{0.545}  && 0.477 & 0.449 \\
                                          &                            & Re  & 1.000* & 0.441  && \textbf{0.603} & 0.515 \\
                                          &                            & F1  & 0.687* & 0.488  && \textbf{0.532} & 0.479 \\
                                          &                            & WAF & 0.359* & \textbf{0.513}  && 0.431 & 0.410 \\ \cmidrule{2-8}
                                          & \multirow{4}{*}{GSE202295} & Pr  & 0.550* & \textbf{1.000}  && 0.533 & 0.538 \\
                                          &                            & Re  & 1.000* & 0.033  && \textbf{0.131} & 0.115 \\
                                          &                            & F1  & 0.709* & 0.063  && \textbf{0.211} & 0.189 \\
                                          &                            & WAF & 0.390* & 0.318  && \textbf{0.381} & 0.372 \\ \midrule
\multirow{12}{*}{\makecell[c]{Coronary \\ Artery \\ Disease}} 
                                          & \multirow{4}{*}{GSE12288}  & Pr  & 0.000* & 0.495* && \textbf{0.463} & 0.462 \\
                                          &                            & Re  & 0.000* & 1.000* && \textbf{0.345} & 0.327 \\
                                          &                            & F1  & 0.000* & 0.663* && \textbf{0.396} & 0.383 \\
                                          &                            & WAF & 0.338* & 0.328* && \textbf{0.468} & 0.466 \\ \cmidrule{2-8}
                                          & \multirow{4}{*}{GSE20681}  & Pr  & 0.500* & 0.500* && 0.532 & \textbf{0.558} \\
                                          &                            & Re  & 1.000* & 1.000* && 0.414 & \textbf{0.485} \\
                                          &                            & F1  & 0.667* & 0.667* && 0.466 & \textbf{0.519} \\
                                          &                            & WAF & 0.333* & 0.333* && 0.519 & \textbf{0.549} \\ \cmidrule{2-8}
                                          & \multirow{4}{*}{GSE42148}  & Pr  & 0.000* & \textbf{0.471}  && 0.455 & 0.400 \\
                                          &                            & Re  & 0.000* & \textbf{0.615}  && 0.385 & 0.462 \\
                                          &                            & F1  & 0.000* & \textbf{0.533}  && 0.417 & 0.429 \\
                                          &                            & WAF & 0.288* & 0.391  && \textbf{0.417} & 0.324 \\ \midrule
\multirow{12}{*}{\makecell[c]{Breast \\ Cancer}}            
                                          & \multirow{4}{*}{GSE9574}   & Pr  & 0.000* & \textbf{1.000}  && 0.500 & 0.464 \\
                                          &                            & Re  & 0.000* & 0.071  && \textbf{1.000} & 0.929 \\
                                          &                            & F1  & 0.000* & 0.133  && \textbf{0.667} & 0.619 \\
                                          &                            & WAF & 0.353* & \textbf{0.425}  && 0.386 & 0.299 \\ \cmidrule{2-8}
                                          & \multirow{4}{*}{GSE10810}  & Pr  & 0.534* & 0.534* && \textbf{0.718} & \textbf{0.718} \\
                                          &                            & Re  & 1.000* & 1.000* && \textbf{0.903} & \textbf{0.903} \\
                                          &                            & F1  & 0.697* & 0.697* && \textbf{0.800} & \textbf{0.800} \\
                                          &                            & WAF & 0.372* & 0.372* && \textbf{0.751} & \textbf{0.751} \\ \cmidrule{2-8}
                                          & \multirow{4}{*}{GSE86374}  & Pr  & 0.000* & 0.780* && \textbf{0.947} & \textbf{0.947} \\
                                          &                            & Re  & 0.000* & 1.000* && \textbf{0.573} & \textbf{0.573} \\
                                          &                            & F1  & 0.000* & 0.876* && \textbf{0.714} & \textbf{0.714} \\
                                          &                            & WAF & 0.079* & 0.683* && \textbf{0.671} & \textbf{0.671} \\ \bottomrule

\end{tabular}
\end{adjustbox}
\vspace{-0.4cm}
\end{table}

Table~\ref{tab:performance2} presents the performance metrics for the transfer learning setting. Unlike the single-dataset and multi-dataset learning settings, where different test sets are used, the entire dataset serves as the test set. As a result, each dataset has a single performance value.
The transfer learning setting is very interesting because it reflects the real-world challenge of generalizing models across diverse clinical settings, enabling predictions on unseen patient data. 
It is important to highlight that when using gene expression values to represent patients, the transfer learning setting is impossible if there are no genes in common between datasets.
Even in the baseline variation which includes all genes, this leads to a scenario where any non-zero features present in the training set end up aligning with zero values in the test set.
Although this is not true for our datasets since there is considerable gene overlap between datasets (Figure~\ref{fig:diagramvenn}), the transfer learning setting using gene expression values is not successful in most datasets.
For the baseline using gene expression data from all genes, it predicts the same label for all instances in the test set in 8 out of 9 datasets. The baseline using overlapping genes demonstrates better performance; however, in 4 datasets, it still exhibits the same issue. In two datasets, while most instances are predicted as 1, resulting in a recall of 1, not all instances are classified with this label.
In contrast, our methodology does not result in any classifier assigning the same label to all instances. 
Notably, in the breast cancer datasets (GSE10810 and GSE86374), our methodology in the transfer learning setting achieves performance results similar to those obtained in single and multi-dataset settings.

When comparing the two variants of methodology, one employing MLP and the other employing GCN, the results indicate that the best ML model depends on the setting. In the single dataset learning setting, GCN demonstrates superior weighted average F1-score values over MLP in five out of nine datasets, showcasing its efficacy in capturing complex relationships within data. However, in the multi-dataset learning setting, MLP outperforms GCN in six datasets. MLP seems to maintain its advantage over GCN in the transfer learning setting, exhibiting superior performance in seven out of nine datasets. This could be attributed to the GCN challenge of overfitting~\cite{Rong2020DropEdge}, as well as to the relevant subgraphs being larger (and, thus, interesting relations spanning more hops) in the multi-dataset and transfer learning settings.

\begin{figure}[t!]\centering
\subfloat[Using gene expression values for all genes.]{\includegraphics[width=0.45\linewidth]{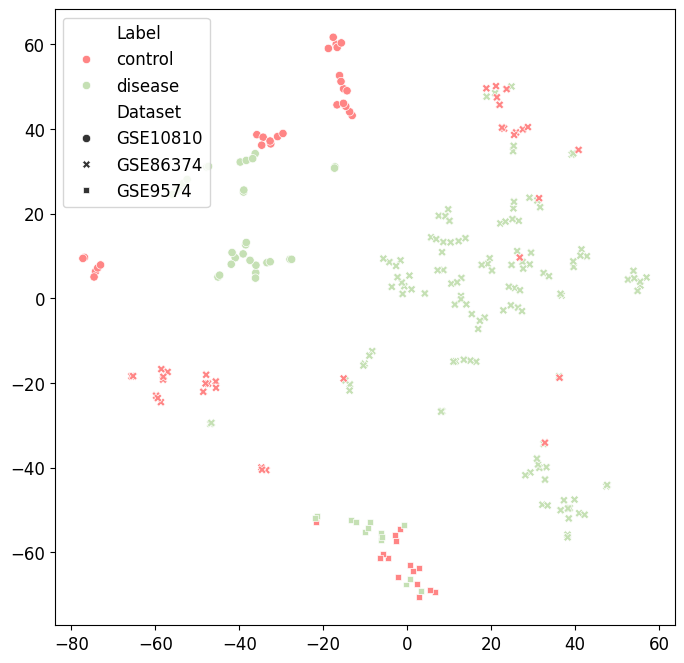}}\hfill
\subfloat[Using gene expression values for overlapping genes.]{\includegraphics[width=0.45\linewidth]{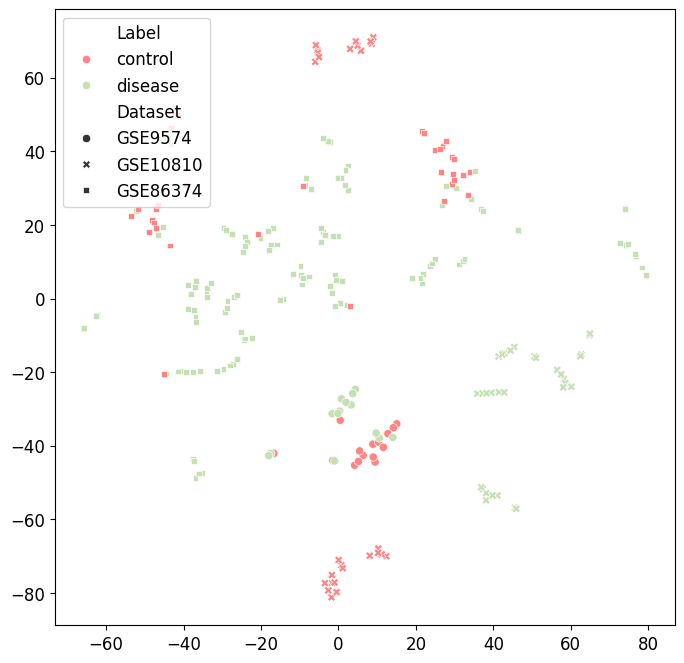}}\hfill
\subfloat[Using KG embeddings.]{\includegraphics[width=0.45\linewidth]{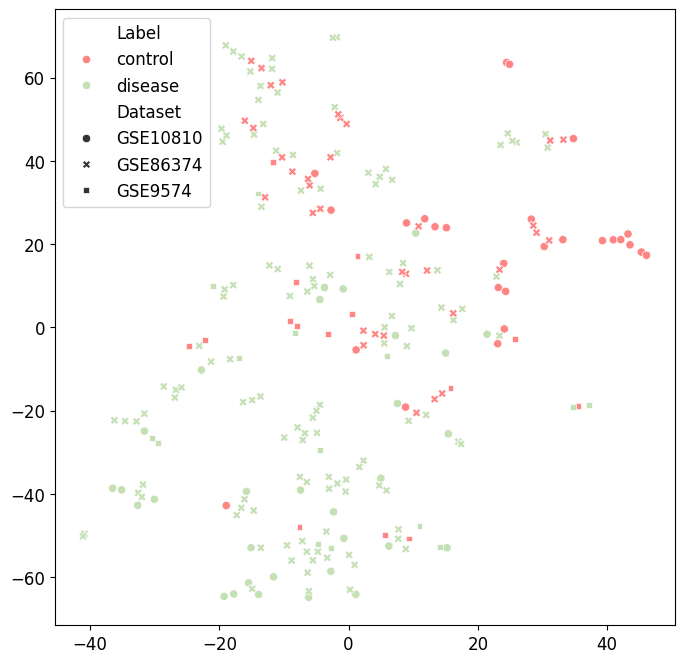}}\hfill 
\caption{t-SNE plots comparing patient representations based on the gene expression values (using all genes or only the overlapping genes across the three datasets) to patient representations generated based on KG embeddings. Each point represents a patient, with the color indicating the label and the shape indicating the dataset they originate from.}
\label{fig:tsne}
\end{figure}

To better understand the differences between the patient representations using the gene expression values and the ones obtained using our KG embeddings, we plot all the patient representations using t-SNE~\cite{van2008visualizing}, a statistical method for visualizing high-dimensional data. Figure~\ref{fig:tsne} shows the representations using gene expression values and our methodology that employs KG embedding methods for breast cancer datasets. By comparing the plots, we observe that our methodology represents patients from different datasets within the same semantic space. In contrast, when using gene expression values, we can distinctly identify three clusters, each corresponding to a different dataset. This illustrates how the multi-dataset and transfer learning cases can benefit from the methodology and combine data from different datasets. 

\begin{figure}[t!]
    \centering
    \includegraphics[width=0.7\linewidth]{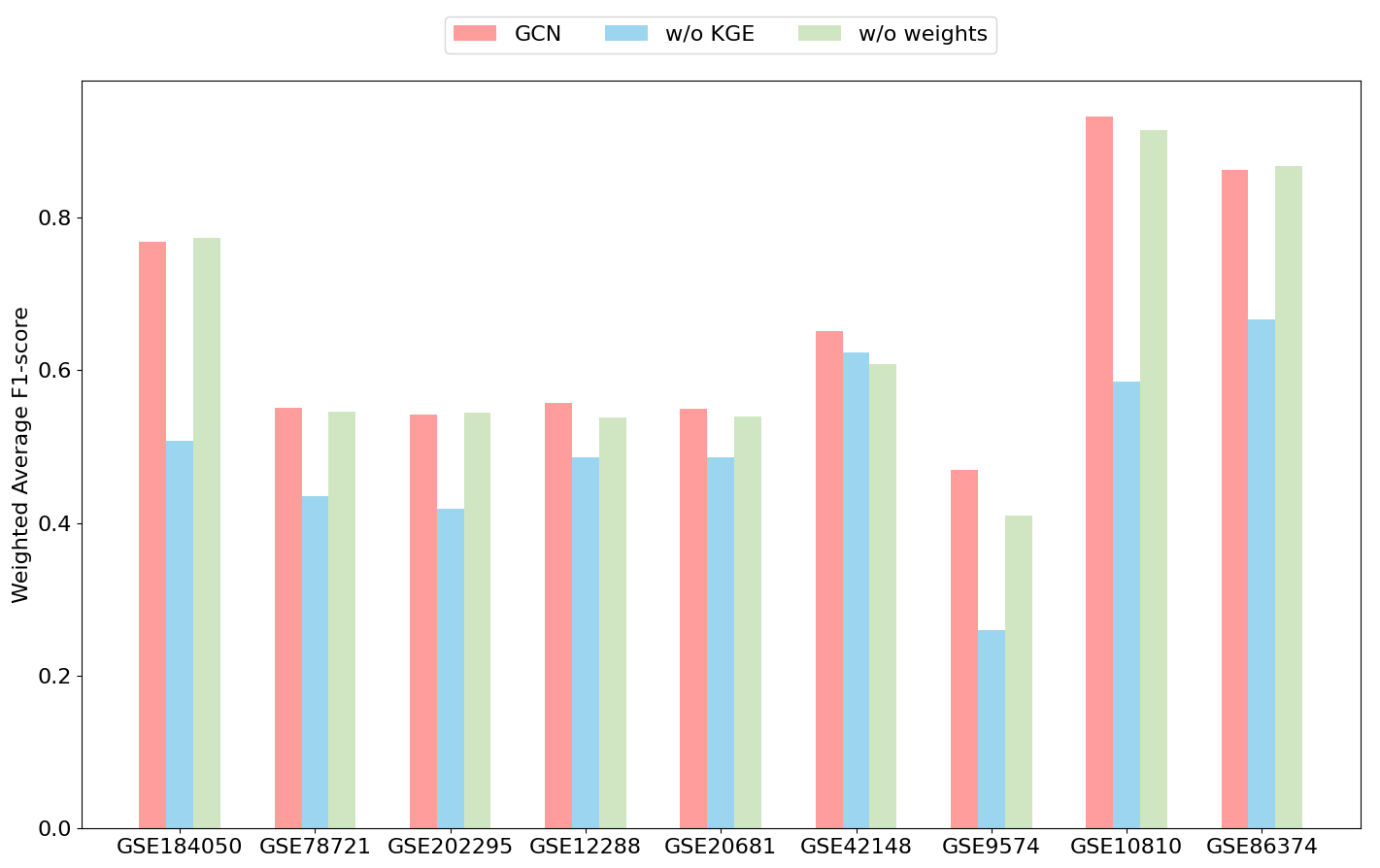}
    \caption{Bar plot depicting the weighted average F-score (WAF) comparisons between different GCN configurations: one using weighted edges and KG embeddings as node features (pink bars), another with randomly initialized node features (blue bars), and another without weighted edges (green bars). No hyperparameter optimization was performed, so all variations used the same setting.}
    \label{fig:ablation-studies}
\end{figure}

In order to validate the effectiveness of each component of GCN architecture, we perform ablation studies by evaluating the performance of a GCN when the input node features are replaced with randomly initialized values and when the model receives as input unweighted graph.
Figure~\ref{fig:ablation-studies} provides a comprehensive analysis of the performance variations in terms of weighted average F1-score when using randomly initialized features versus KG embeddings as node features and weighted edges versus unweighted edges.
The results demonstrate that integrating KG embeddings with GCNs consistently outperforms GCNs lacking node features across all datasets. This highlights the significant impact of node features and confirms that KG embeddings effectively generate meaningful numerical features that enhance patient diagnosis.
The influence of using weighted edges appears to be less significant. In most datasets, the differences are minimal, and in some cases, the unweighted graph achieves better results. These variations can be attributed to the fact that we only have edges between patients and genes when the z-score is greater than 1. Consequently, the variations in edge weights might be too small to convey additional information beyond the presence of an edge.

\section{Conclusion}

Several ML-based approaches for patient diagnosis rely on analyzing expression data, which offers a detailed molecular profile reflecting gene activity and regulation, thereby revealing relationships between specific genes and diseases' pathogenesis. However, exploring expression data in ML tasks, such as patient diagnosis, poses challenges due to the very low number of patients in existing gene expression datasets and the lack of data integration. 

We present a methodology that allows gene expression data from several datasets to be comprehensively represented within a KG. This KG is able to integrate distinct datasets from different experimental studies into a single, unified space using domain-specific knowledge. 
The KG is then exploited by KG embedding methods or, alternatively, GCNs to support patient diagnosis. We investigated the impact of integrating domain-specific knowledge and multiple gene expression datasets and the efficacy of transferring a model trained on one dataset to another for testing. The proposed methodology is versatile and can be extended to combining datasets with incompatible features beyond the gene expression domain, offering a tool for integrating diverse data types.
Future work could explore additional domain-specific knowledge for gene expression datasets and incorporate additional performance metrics to gain deeper insights and broaden the approach's applicability.
%
%
%
\bibliographystyle{splncs04}
\bibliography{mybibliography}

\clearpage
\appendix

\section{Hyperparameters}

The parameters used for the RDF2vec model are described in Table~\ref{tab:rdf2vec-parameters}.
To run supervised learning methods, we optimized certain parameters. The parameters are supplied in Tables~\ref{tab:parameters-gcn} and \ref{tab:parameters-mlp}.

\begin{table}[h!]
\scriptsize
\caption{Parameters for RDF2vec embedding method.}
\label{tab:rdf2vec-parameters}
\begin{tabular}{ll}
\toprule
\textbf{Parameter} &  \textbf{Value} \\ 
\midrule
Embedding size & 500  \\ 
Maximum \# of walks per entity & 500 \\ 
Maximum walk depth & 4 \\ 
Word2vec model & skip-gram \\ 
Epochs for Word2vec & 5 \\ 
Window for Word2vec & 5 \\ 
Minimum count for Word2vec & 1 \\ 
Optimization for softmax &  negative sampling \\ 
Noise words drawn in neg. sampling & 5 \\ 
Learning rate & 0.025 \\
\bottomrule
\end{tabular}
\end{table}


\begin{table}[h!]
\scriptsize
\caption{MLP parameters that have been optimized.}
\label{tab:parameters-mlp}
\begin{tabular}{ll}
\toprule
\textbf{Parameter} & \textbf{Values}\\ \midrule
Hidden layer sizes & (100,), (50,50), (30, 20, 10) \\
Activation function & hyperbolic tan function, relu function \\
Optimization method & SGD, Adam  \\
Alpha & 0.0001, 0.001, 0.01 \\
Learning rate  & constant, adaptive  \\
\bottomrule
\end{tabular}
\end{table}


\begin{table}[h!]
\scriptsize
\caption{GCN parameters that have been optimized.}
\label{tab:parameters-gcn}
\begin{tabular}{ll}
\hline
\textbf{Parameter} & \textbf{Values}\\ \hline
In channels & 500 \\
Out channels & 2 \\ 
Hidden channels & 16, 32 \\ 
Number of conv layers & 2, 3, 5, 6  \\
Learning rate & 0.001, 0.01, 0.1, 0.2 \\
Dropout & 0.2, 0.3, 0.5 \\
Aggregation &  Avg, Max \\
Epochs & 1000 \\
\bottomrule
\end{tabular}
\end{table}

\end{document}